\documentclass[11pt,a4paper]{article}
\usepackage[hyperref]{acl2020}
\usepackage{times}
\usepackage{latexsym}

\usepackage{amsmath,amsfonts,bm}

\def\eqref#1{equation~\ref{#1}}

\def\1{\bm{1}}

\def\vb{{\bm{b}}}

\def\vf{{\bm{f}}}

\def\vv{{\bm{v}}}

\def\vx{{\bm{x}}}

\def\vz{{\bm{z}}}

\def\mW{{\bm{W}}}

\DeclareMathAlphabet{\mathsfit}{\encodingdefault}{\sfdefault}{m}{sl}
\SetMathAlphabet{\mathsfit}{bold}{\encodingdefault}{\sfdefault}{bx}{n}

\newcommand{\R}{\mathbb{R}}

\newcommand{\BERT}{\mathrm{BERT}}

\DeclareMathOperator*{\argmax}{arg\,max}

\usepackage{graphicx}
\usepackage{algorithm}
\usepackage{algpseudocode}

\aclfinalcopy 
\def\aclpaperid{2117} 

\newcommand{\BLEURT}{\textsc{Bleurt}}

\newcommand{\BLEURTsys}{\texttt{BLEURT}}
\newcommand{\BLEURTbase}{\texttt{BLEURTbase}}
\newcommand{\BLEURTnopre}{\texttt{BLEURT -pre}}
\newcommand{\BLEURTbasenopre}{\texttt{BLEURTbase -pre}}
\newcommand{\BLEURTnoprenowmt}{\texttt{BLEURT -pre -wmt}}
\newcommand{\BLEURTnowmt}{\texttt{BLEURT -wmt}}

\newcommand{\iid}{\textsc{iid}}

\newcommand{\en}{\texttt{en}}
\newcommand{\de}{\texttt{de}}
\newcommand{\fr}{\texttt{fr}}

\newcommand{\CLS}{\textrm{[CLS]}}

\newcommand{\arkcomment}[3]{\ensuretext{\textcolor{#3}{[#1 #2]}}}
\renewcommand{\arkcomment}[3]{} 

\title{BLEURT: Learning Robust Metrics for Text Generation}

\author{Thibault Sellam \ \ \ Dipanjan Das \ \ \ Ankur P. Parikh \\
Google Research \\
New York, NY \\
\texttt{\{tsellam, dipanjand, aparikh \}@google.com}
}

\begin{document}

\maketitle

\begin{abstract}
Text generation has made significant advances in the last few years. Yet, evaluation metrics have lagged behind, as the most popular choices (e.g., BLEU and ROUGE) may correlate poorly with human judgments. We propose \BLEURT, a learned evaluation metric based on BERT that can model human judgments with a few thousand possibly biased training examples. A key aspect of our approach is a novel pre-training scheme that uses millions of synthetic examples to help the model generalize. \BLEURT{} provides state-of-the-art results on the last three years of the WMT Metrics shared task and the WebNLG Competition dataset. In contrast to a vanilla BERT-based approach, it yields superior results even when the training data is scarce and out-of-distribution.
\end{abstract}

\section{Introduction}

In the last few years, research in natural text generation (NLG) has made significant progress, driven largely by the neural encoder-decoder paradigm~\citep{sutskever2014sequence,bahdanau2014neural} which can tackle a wide array of tasks including translation~\citep{koehn2009statistical}, summarization~\citep{mani1999advances,chopra2016abstractive}, structured-data-to-text generation~\cite{mckeown1992text,kukich1983design,wiseman2017challenges}  dialog~\cite{smith1994spoken,vinyals2015neural} and image captioning~\cite{fang2015captions}. However, progress is increasingly impeded by the shortcomings of existing metrics~\citep{wiseman2017challenges,ma2019results,tian2019sticking}. 

Human evaluation is often the best indicator of the quality of a system. However, designing crowd sourcing experiments is an expensive and high-latency process, which does not easily fit in a daily model development pipeline. Therefore, NLG researchers commonly use \emph{automatic evaluation metrics}, which provide an acceptable proxy for quality and are very cheap to compute. This paper investigates sentence-level, reference-based metrics, which describe the extent to which a \emph{candidate} sentence is similar to a \emph{reference} one. The exact definition of similarity may range from string overlap to logical entailment.

The first generation of metrics relied on handcrafted rules that measure the surface similarity between the sentences. To illustrate, BLEU~\citep{papineni2002bleu} and ROUGE~\citep{lin2004rouge}, two  popular metrics, rely on N-gram overlap. Because those metrics are only sensitive to lexical variation, they cannot appropriately reward semantic or syntactic variations of a given reference. Thus, they have been repeatedly shown to correlate poorly with human judgment, in particular when all the systems to compare have a similar level of accuracy~\cite{liu2016not, novikova2017we,chaganty2018price}. 

Increasingly, NLG researchers have addressed those problems  by injecting \emph{learned} components in their metrics. To illustrate, consider the WMT Metrics Shared Task, an annual benchmark in which translation metrics are compared on their ability to imitate human assessments. The last two years  of the competition were largely dominated by neural net-based approaches, RUSE, YiSi and ESIM~\cite{ma2018results,ma2019results}. Current approaches largely fall into two categories. \emph{Fully learned metrics}, such as BEER, RUSE, and ESIM are trained end-to-end, and they typically rely on handcrafted features and/or learned embeddings. Conversely, \emph{hybrid metrics}, such as YiSi and BERTscore combine trained elements, e.g., contextual embeddings, with handwritten logic, e.g., as token alignment rules. The first category typically offers great expressivity: if a training set of human ratings data is available, the metrics may take full advantage of it and fit the ratings distribution tightly. Furthermore, learned metrics can be tuned to measure task-specific properties, such as fluency, faithfulness, grammar, or style. On the other hand, hybrid metrics offer robustness. They may provide better results when there is little to no training data, and they do not rely on the assumption that training and test data are identically distributed.

And indeed, the \iid{} assumption is particularly problematic in NLG evaluation because of \emph{domain drifts}, that have been the main target of the metrics literature, but also because of \emph{quality drifts}: NLG systems tend to get better over time, and therefore a model trained on ratings data from 2015 may fail to distinguish top performing systems in 2019, especially for newer research tasks. An ideal learned metric would be able to both take full advantage of available ratings data for training, and be robust to distribution drifts, i.e., it should be able to \emph{extrapolate}.

Our insight is that it is possible to combine expressivity and robustness by pre-training a fully learned metric on large amounts of synthetic data, before fine-tuning it on human ratings. To this end, we introduce \BLEURT,\footnote{Bilingual Evaluation Understudy with Representations from Transformers. We refer the intrigued reader to \citealt{papineni2002bleu} for a justification of the term \emph{understudy}.}  a text generation metric based on BERT~\cite{devlin2018bert}. A key ingredient of \BLEURT~is a novel pre-training scheme, which uses random perturbations of Wikipedia sentences augmented with a diverse set of lexical and semantic-level supervision signals. 



To demonstrate our approach, we train \BLEURT{} for English and evaluate it under different generalization regimes. We first verify that it provides state-of-the-art results on all recent years of the WMT Metrics Shared task (2017 to 2019, to-English language pairs). We then stress-test its ability to cope with quality drifts with a synthetic benchmark based on WMT 2017. Finally, we show that it can easily adapt to a different domain with three tasks from a data-to-text dataset, WebNLG 2017~\cite{gardent2017webnlg}. Ablations show that our synthetic pretraining scheme increases performance in the \iid{} setting, and is critical to ensure robustness when the training data is scarce, skewed, or out-of-domain.

The code and pre-trained models are available online\footnote{\url{http://github.com/google-research/bleurt}}.

\section{Preliminaries}
Define $\vx = (x_1,..,x_{r})$ to be the reference sentence of length $r$ where each $x_i$ is a token and let $\tilde{\vx} = (\tilde{x}_1,..,\tilde{x}_{p})$ be a prediction sentence of length $p$. Let $\{(\vx_i, \tilde{\vx}_i, y_i)\}_{n=1}^{N}$ be a training dataset of size $N$ where $y_i \in \R$ is the human rating that indicates how good $\tilde{\vx}_i$ is with respect to $\vx_i$. Given the training data, our goal is to learn a function $\vf : (\vx, \tilde{\vx}) \rightarrow y$ that predicts the human rating.

\section{Fine-Tuning BERT for Quality Evaluation}
\label{sec:fine-tuning}
Given the small amounts of rating data available, it is natural to leverage unsupervised representations for this task. In our model, we use BERT (Bidirectional Encoder Representations from
Transformers)~\citep{devlin2018bert}, which is an unsupervised technique that learns contextualized representations of sequences of text. Given $\vx$ and $\tilde{\vx}$, BERT is a Transformer~\citep{vaswani2017attention} that returns a sequence of contextualized vectors:
$$
\vv_{\mathrm{[CLS]}}, \vv_{x_1},...,\vv_{x_r}, \vv_1,...,\vv_{\tilde{x}_p} = \BERT(\vx, \tilde{\vx})
$$
where $\vv_{\mathrm{[CLS]}}$ is the representation for the special $\mathrm{[CLS]}$ token. As described by~\newcite{devlin2018bert}, we add a linear layer on top of the $\mathrm{[CLS]}$ vector to predict the rating: 
$$
\hat{y} = \vf(\vx, \tilde{\vx}) = \mW \tilde{\vv}_{\mathrm{[CLS]}} + \vb 
$$
where $\mW$ and $\vb$ are the weight matrix and bias vector respectively. Both the above linear layer as well as the BERT parameters are trained (i.e. fine-tuned) on the supervised data which typically numbers in a few thousand examples. We use the regression loss $\ell_{\textrm{supervised}} = \frac{1}{N} \sum_{n=1}^{N} \|y_i  - \hat{y} \|^2 $.

Although this approach is quite straightforward, we will show in Section~\ref{sec:experiments} that it gives state-of-the-art results on WMT Metrics Shared Task 17-19, which makes it a high-performing evaluation metric. However, fine-tuning BERT requires a sizable amount of \iid{} data, which is less than ideal for a metric that should generalize to a variety of tasks and model drift.

\section{Pre-Training on Synthetic Data}
\label{sec:pretraining}

The key aspect of our approach is a pre-training technique that we use to ``warm up'' BERT before fine-tuning on rating data.\footnote{To clarify, our pre-training scheme is an addition, not a replacement to BERT's initial training~\cite{devlin2018bert} and happens after it.} We generate a large number of of synthetic reference-candidate pairs $(\vz, \tilde{\vz})$, and we train BERT on several lexical- and semantic-level supervision signals with a multitask loss. As our experiments will show, \BLEURT{} generalizes much better after this phase, especially with incomplete training data.

Any pre-training approach requires a dataset and a set of pre-training tasks. Ideally, the setup should resemble the final NLG evaluation task, i.e., the sentence pairs should be distributed similarly and the pre-training signals should correlate with human ratings. Unfortunately, we cannot have access to the NLG models that we will evaluate in the future. Therefore, we optimized our scheme for generality, with three requirements. (1)~The set of reference sentences should be large and diverse, so that \BLEURT{} can cope with a wide range of NLG domains and tasks. (2)~The sentence pairs should contain a wide variety of lexical, syntactic, and semantic dissimilarities. The aim here is to anticipate all variations that an NLG system may produce, e.g., phrase substitution, paraphrases, noise, or omissions. (3)~The pre-training objectives should effectively capture those phenomena, so that \BLEURT{} can learn to identify them. The following sections present our approach.

\begin{table*}[!t]
\begin{small}
\begin{center}
  \begin{tabular}{ l c c }
    \hline
    Task Type & Pre-training Signals & Loss Type \\ 
    \hline
    BLEU & $ \bm{\tau}_{\text{BLEU}}$ & Regression \\
    ROUGE & $\bm{\tau}_{\text{ROUGE}} = (\tau_{\text{ROUGE-P}}, \tau_{\text{ROUGE-R}}, \tau_{\text{ROUGE-F}}) $ & Regression \\
    BERTscore & $\bm{\tau}_{\text{BERTscore}} = (\tau_{\text{BERTscore-P}}, \tau_{\text{BERTscore-R}}, \tau_{\text{BERTscore-F}}) $ & Regression \\
    Backtrans. likelihood & $\bm{\tau}_{\text{en-fr}, \vz \mid \tilde{\vz}}$, $\bm{\tau}_{\text{en-fr}, \tilde{\vz} \mid \vz}$, $\bm{\tau}_{\text{en-de}, \vz \mid \tilde{\vz}}$, $\bm{\tau}_{\text{en-de}, \tilde{\vz} \mid \vz}$  & Regression \\
    Entailment & $\bm{\tau}_{\text{entail}} = (\tau_{\text{Entail}}, \tau_{\text{Contradict}}, \tau_{\text{Neutral}})$ & Multiclass\\
    Backtrans. flag & $\bm{\tau}_{\text{backtran\_flag}}$ & Multiclass \\
    \hline
  \end{tabular}
\end{center}
\end{small}
\caption{Our pre-training signals.}
\label{tab:pre-training-signals}
\end{table*}

\subsection{Generating Sentence Pairs}
\label{subsec:synthetic}


One way to expose \BLEURT{} to a wide variety of sentence differences is to use existing sentence pairs datasets~\citep{bowman2015large,williams2017broad,wang2018glue}. These sets are a rich source of related sentences, but they may fail to capture the errors and alterations that NLG systems produce (e.g., omissions, repetitions, nonsensical substitutions).  We opted for an automatic approach instead, that can be scaled arbitrarily and at little cost: we generate synthetic sentence pairs $(\vz, \tilde{\vz})$ by randomly perturbing 1.8 million segments $\vz$ from Wikipedia. We use three techniques: mask-filling with BERT, backtranslation, and randomly dropping out words. We obtain about 6.5 million perturbations $\tilde{\vz}$. Let us describe those techniques.


\paragraph{Mask-filling with BERT:} BERT's initial training task is to fill gaps (i.e., masked tokens) in tokenized sentences. We leverage this functionality by inserting masks at random positions in the Wikipedia sentences, and fill them with the language model. Thus, we introduce lexical alterations while maintaining the fluency of the sentence. We use two masking strategies---we either introduce the masks at random positions in the sentences, or we create contiguous sequences of masked tokens. More details are provided in the Appendix.

\paragraph{Backtranslation:} We generate paraphrases and perturbations with backtranslation, that is, round trips from English to another language and then back to English with a translation model~\citep{bannard2005paraphrasing,ganitkevitch2013ppdb,sennrich2015improving}. Our primary aim is to create variants of the reference sentence that preserves semantics. Additionally, we use the mispredictions of the backtranslation models as a source of realistic alterations.

\paragraph{Dropping words:} We found it useful in our experiments to randomly drop words from the synthetic examples above to create other examples. This method prepares \BLEURT{} for ``pathological'' behaviors or NLG systems, e.g., void predictions, or sentence truncation.

\subsection{Pre-Training Signals}
\label{subsec:signals}

The next step is to augment each sentence pair $(\vz, \tilde{\vz})$ with a set of pre-training signals $\{\bm{\tau}_k\}$, where $\bm{\tau}_k$ is the target vector of pre-training task~$k$. Good pre-training signals should capture a wide variety of lexical and semantic differences. They should also be cheap to obtain, so that the approach can scale to large amounts of synthetic data. The following section presents our 9 pre-training tasks, summarized in Table~\ref{tab:pre-training-signals}. Additional implementation details are in the Appendix.

\paragraph{Automatic Metrics:} We create three signals $\bm{\tau_{\text{BLEU}}}$, $\bm{\tau_{\text{ROUGE}}}$, and $\bm{\tau_{\text{BERTscore}}}$ with sentence BLEU ~\citep{papineni2002bleu}, ROUGE ~\citep{lin2004rouge}, and BERTscore~\citep{zhang2019bertscore} respectively (we use precision, recall and F-score for the latter two).

\paragraph{Backtranslation Likelihood:}
The idea behind this signal is to leverage existing translation models to measure semantic equivalence.
Given a pair $(\vz, \tilde{\vz})$, this training signal measures the probability that $\tilde{\vz}$ is a backtranslation of $\vz$, $P(\tilde{\vz} | \vz)$, normalized by the length of $\tilde{\vz}$. Let $P_{\en \rightarrow \fr}(\vz_{\fr} | \vz)$ be a translation model that assigns probabilities to French sentences $\vz_{\fr}$ conditioned on English sentences $\vz$ and let $P_{\fr \rightarrow \en}(\vz | \vz_{\fr})$ be a translation model that assigns probabilities to English sentences given french sentences. If $|\tilde{\vz}|$ is the number of tokens in $\tilde{\vz}$,
we define our score as $ \bm{\tau}_{\text{en-fr}, \tilde{\vz} \mid \vz} = \frac{\log P(\tilde{\vz} | \vz)}{|\tilde{\vz}|}$, with:
$$
P(\tilde{\vz} | \vz) =\sum_{\vz_{\fr}} P_{\fr \rightarrow \en}( \tilde{\vz} | \vz_{\fr}) P_{\en \rightarrow \fr} (\vz_{\fr} | \vz)
$$ 
Because computing the summation over all possible French sentences is intractable, we approximate the sum using $\vz_{\fr}^\ast = \argmax P_{\en \rightarrow \fr} (\vz_{\fr} | \vz)$ and we assume that $P_{\en \rightarrow \fr}(\vz_{\fr}^\ast | \vz) \approx 1$:
$$
P(\tilde{\vz} | \vz) \approx  P_{\fr \rightarrow \en}(\tilde{\vz} | \vz_{\fr}^\ast) 
$$
We can trivially reverse the procedure to compute $P(\vz | \tilde{\vz})$, thus we create 4 pre-training signals $\bm{\tau}_{\text{en-fr}, \vz \mid \tilde{\vz}}$, $\bm{\tau}_{\text{en-fr}, \tilde{\vz} \mid \vz}$, $\bm{\tau}_{\text{en-de}, \vz \mid \tilde{\vz}}$, $\bm{\tau}_{\text{en-de}, \tilde{\vz} \mid \vz}$ with two pairs of languages ($\en \leftrightarrow \de$ and $\en \leftrightarrow \fr$) in both directions.

\paragraph{Textual Entailment:} The signal $\bm{\tau}_\text{entail}$ expresses whether $\vz$ entails or contradicts $\tilde{\vz}$ using a classifier. We report the probability of three labels: \textit{Entail}, \textit{Contradict}, and \textit{Neutral}, using BERT fine-tuned on an entailment dataset, MNLI~\cite{devlin2018bert, williams2017broad}.

\paragraph{Backtranslation flag:} The signal $\bm{\tau}_\text{backtran\_flag}$ is a Boolean that indicates whether the perturbation was generated with backtranslation or with mask-filling.

\subsection{Modeling}
\label{sec:modeling}

For each pre-training task, our model uses either a regression or a classification loss. We then aggregate the task-level losses with a weighted sum.

Let $\bm{\tau}_k$ describe the target vector for each task, e.g., the probabilities for the classes \emph{Entail}, \emph{Contradict}, \emph{Neutral}, or the precision, recall, and F-score for ROUGE. If $\bm{\tau}_k$ is a regression task, then the loss used is the $\ell_2$ loss i.e. $\ell_k = \| \bm{\tau}_k - \hat{\bm{\tau}}_k \|_2^2 / |\bm{\tau}_k|$ where $|\bm{\tau}_k|$ is the dimension of $\bm{\tau}_k$ and $\hat{\bm{\tau}}_k$ is computed by using a task-specific linear layer on top of the $\CLS$ embedding: $\hat{\bm{\tau}}_k = \mW_{\tau_k} \tilde{\vv}_{\CLS} + \vb_{\tau_k}$. If  $\bm{\tau}_k$ is a classification task, we use a separate linear layer to predict a logit for each class $c$: $\hat{\bm{\tau}}_{kc} = \mW_{\tau_{kc}} \tilde{\vv}_{\CLS} + \vb_{\tau_{kc}}$, and we use the multiclass cross-entropy loss. We define our aggregate pre-training loss function as  follows:
 \begin{align}
  \ell_{\textrm{pre-training}} = \frac{1}{M} \sum_{m=1}^{M} \sum_{k=1}^{K} \gamma_k \ell_k(\bm{\tau}_k^m, \hat{\bm{\tau}}_k^m)
 \end{align}
 where $\bm{\tau}_k^m$ is the target vector for example $m$, $M$ is number of synthetic examples, and $\gamma_k$ are hyperparameter weights obtained with grid search (more details in the Appendix).

\begin{table*}[ht]
\centering
\scriptsize
\begin{tabular}{lcccccccc}
  \hline
model & cs-en & de-en & fi-en & lv-en & ru-en & tr-en & zh-en & \textbf{avg} \\ 
& $\tau$ / $r$ & $\tau$ / $r$ & $\tau$ / $r$ & $\tau$ / $r$ & $\tau$ / $r$ & $\tau$ / $r$ & $\tau$ / $r$ & $\tau$ / $r$ \\
  \hline
  sentBLEU & 29.6 / 43.2 & 28.9 / 42.2 & 38.6 / 56.0 & 23.9 / 38.2 & 34.3 / 47.7 & 34.3 / 54.0 & 37.4 / 51.3 & 32.4 / 47.5 \\
  MoverScore & 47.6 / 67.0 & 51.2 / 70.8 & NA & NA & 53.4 / 73.8 & 56.1 / 76.2 & 53.1 / 74.4 & 52.3 / 72.4\\
  BERTscore w/ BERT & 48.0 / 66.6 & 50.3 / 70.1 & 61.4 / 81.4 & 51.6 / 72.3 & 53.7 / 73.0 & 55.6 / 76.0 & 52.2 / 73.1 & 53.3 / 73.2 \\ 
  BERTscore w/ roBERTa & 54.2 / 72.6 & 56.9 / 76.0 & 64.8 / 83.2 & 56.2 / 75.7 & 57.2 / 75.2 & 57.9 / 76.1 & 58.8 / 78.9 & 58.0 / 76.8 \\ 
  \hline
  chrF++ & 35.0 / 52.3 & 36.5 / 53.4 & 47.5 / 67.8 & 33.3 / 52.0 & 41.5 / 58.8 & 43.2 / 61.4 & 40.5 / 59.3 & 39.6 / 57.9 \\ 
  BEER & 34.0 / 51.1 & 36.1 / 53.0 & 48.3 / 68.1 & 32.8 / 51.5 & 40.2 / 57.7 & 42.8 / 60.0 & 39.5 / 58.2 & 39.1 / 57.1 \\ 
  \hline
  BLEURTbase -pre & 51.5 / 68.2 & 52.0 / 70.7 & 66.6 / 85.1 & 60.8 / 80.5 & 57.5 / 77.7 & 56.9 / 76.0 & 52.1 / 72.1 & 56.8 / 75.8 \\
  BLEURTbase & 55.7 / 73.4 & 56.3 / 75.7 & 68.0 / 86.8 & \textbf{64.7} / 83.3 & 60.1 / 80.1 & 62.4 / 81.7 & 59.5 / 80.5 &  61.0 / 80.2 \\ 
  \hline
  BLEURT -pre & 56.0 / 74.7 & 57.1 / 75.7 & 67.2 / 86.1 & 62.3 / 81.7 & 58.4 / 78.3 & 61.6 / 81.4 & 55.9 / 76.5 & 59.8 / 79.2 \\ 
  BLEURT & \textbf{59.3} / \textbf{77.3} & \textbf{59.9} / \textbf{79.2} & \textbf{69.5} / \textbf{87.8} & 64.4 / \textbf{83.5} & \textbf{61.3} / \textbf{81.1} & \textbf{62.9} / \textbf{82.4} & \textbf{60.2} / \textbf{81.4} & \textbf{62.5} / \textbf{81.8} \\
   \hline
\end{tabular}
\caption{Agreement with human ratings on the WMT17 Metrics Shared Task. The metrics are Kendall Tau ($\tau$) and the Pearson correlation ($r$, the official metric of the shared task), divided by 100.}
\label{table:wmt17}
\end{table*}

\begin{table*}[ht]
\centering
\scriptsize
\begin{tabular}{lcccccccc}
  \hline
model & cs-en & de-en & et-en & fi-en & ru-en & tr-en & zh-en & \textbf{avg} \\ 
& $\tau$ / DA & $\tau$ / DA & $\tau$ / DA & $\tau$ / DA & $\tau$ / DA & $\tau$ / DA & $\tau$ / DA & $\tau$ / DA \\
  \hline
  sentBLEU & 20.0 / 22.5 & 31.6 / 41.5 & 26.0 / 28.2 & 17.1 / 15.6 & 20.5 / 22.4 & 22.9 / 13.6 & 21.6 / 17.6 & 22.8 / 23.2  \\ 
  BERTscore w/ BERT & 29.5 / 40.0 & 39.9 / 53.8 & 34.7 / 39.0 & 26.0 / 29.7 & 27.8 / 34.7 & 31.7 / 27.5 & 27.5 / 25.2  & 31.0 / 35.7 \\ 
  BERTscore w/ roBERTa & 31.2 / 41.1 & 42.2 / 55.5 & 37.0 / 40.3 & 27.8 / 30.8 & 30.2 / 35.4 & 32.8 / 30.2 & 29.2 / 26.3 & 32.9 / 37.1 \\ 
  \hline
  Meteor++ & 22.4 / 26.8 & 34.7 / 45.7 & 29.7 / 32.9 & 21.6 / 20.6 & 22.8 / 25.3 & 27.3 / 20.4 & 23.6 / 17.5* & 26.0 / 27.0 \\ 
  RUSE & 27.0 / 34.5 & 36.1 / 49.8 & 32.9 / 36.8 & 25.5 / 27.5 & 25.0 / 31.1 & 29.1 / 25.9 & 24.6 / 21.5* & 28.6 / 32.4 \\ 
  YiSi1 & 23.5 / 31.7 & 35.5 / 48.8 & 30.2 / 35.1 & 21.5 / 23.1 & 23.3 / 30.0 & 26.8 / 23.4 & 23.1 / 20.9 & 26.3 / 30.4 \\ 
  YiSi1 SRL 18 & 23.3 / 31.5 & 34.3 / 48.3 & 29.8 / 34.5 & 21.2 / 23.7 & 22.6 / 30.6 & 26.1 / 23.3 & 22.9 / 20.7 & 25.7 / 30.4 \\ 
  \hline
  BLEURTbase -pre & 33.0 / 39.0 & 41.5 / 54.6 & 38.2 / 39.6 & 30.7 / 31.1 & 30.7 / 34.9 & 32.9 / 29.8 & 28.3 / 25.6 & 33.6 / 36.4 \\ 
  BLEURTbase & 34.5 / \textbf{42.9} & 43.5 / 55.6 & 39.2 / 40.5 & 31.5 / 30.9 & 31.0 / 35.7 & 35.0 / 29.4 & 29.6 / \textbf{26.9} & 34.9 / 37.4 \\ 
  \hline
  BLEURT -pre & 34.5 / 42.1 & 42.7 / 55.4 & 39.2 / 40.6 & 31.4 / 31.6 & 31.4 / 34.2 & 33.4 / 29.3 & 28.9 / 25.6 & 34.5 / 37.0 \\ 
  BLEURT & \textbf{35.6} / 42.3  & \textbf{44.2} / \textbf{56.7} & \textbf{40.0} / \textbf{41.4} & \textbf{32.1} / \textbf{32.5} & \textbf{31.9} / \textbf{36.0} & \textbf{35.5} / \textbf{31.5} & \textbf{29.7} / 26.0 & \textbf{35.6} / \textbf{38.1}  \\ 
   \hline
\end{tabular}
\caption{Agreement with human ratings on the WMT18 Metrics Shared Task. The metrics are Kendall Tau ($\tau$) and WMT's Direct Assessment metrics divided by 100. The star * indicates results that are more than 0.2 percentage points away from the official WMT results (up to 0.4 percentage points away).}.
\label{table:wmt18}
\end{table*}

\begin{table*}[ht]
\centering
\scriptsize
\begin{tabular}{lcccccccc}
  \hline
model & de-en & fi-en & gu-en & kk-en & lt-en & ru-en & zh-en & \textbf{avg} \\ 
& $\tau$ / DA & $\tau$ / DA & $\tau$ / DA & $\tau$ / DA & $\tau$ / DA & $\tau$ / DA & $\tau$ / DA & $\tau$ / DA \\
  \hline
  sentBLEU & 19.4 /  5.4 & 20.6 / 23.3 & 17.3 / 18.9 & 30.0 / 37.6 & 23.8 / 26.2 & 19.4 / 12.4 & 28.7 / 32.2 & 22.7 / 22.3 \\
  BERTscore w/ BERT  & 26.2 / 17.3 & 27.6 / 34.7 & 25.8 / 29.3 & 36.9 / 44.0 & 30.8 / 37.4 & 25.2 / 20.6 & 37.5 / 41.4 & 30.0 / 32.1 \\ 
  BERTscore w/ roBERTa & 29.1 / 19.3 & 29.7 / 35.3 & 27.7 / \textbf{32.4} & 37.1 / 43.1 & 32.6 / 38.2 & 26.3 / \textbf{22.7} & 41.4 / \textbf{43.8} & 32.0 / \textbf{33.6} \\ 
  \hline
  ESIM & 28.4 / 16.6 & 28.9 / 33.7 & 27.1 / 30.4 & 38.4 / 43.3 & 33.2 / 35.9 & 26.6 / 19.9 & 38.7 / 39.6 & 31.6 / 31.3 \\ 
  YiSi1 SRL 19 & 26.3 / \textbf{19.8} & 27.8 / 34.6 & 26.6 / 30.6 & 36.9 / 44.1 & 30.9 / 38.0 & 25.3 / 22.0 & 38.9 / 43.1 & 30.4 / 33.2 \\ 
  \hline
  BLEURTbase -pre & 30.1 / 15.8 & 30.4 / 35.4 & 26.8 / 29.7 & 37.8 / 41.8 & 34.2 / 39.0 & 27.0 / 20.7 & 40.1 / 39.8 & 32.3 / 31.7 \\ 
  BLEURTbase & 31.0 / 16.6 & 31.3 / 36.2 & 27.9 / 30.6 & \textbf{39.5} / \textbf{44.6} & \textbf{35.2} / 39.4 & \textbf{28.5} / 21.5 & 41.7 / 41.6 & 33.6 / 32.9 \\ 
  \hline
  BLEURT -pre & 31.1 / 16.9 & 31.3 / \textbf{36.5} & 27.6 / 31.3 & 38.4 / 42.8 & 35.0 / 40.0 & 27.5 / 21.4 & 41.6 / 41.4 & 33.2 / 32.9 \\ 
  BLEURT & \textbf{31.2} / 16.9 & \textbf{31.7} / 36.3 & \textbf{28.3} / 31.9 & \textbf{39.5} / \textbf{44.6} & \textbf{35.2} / \textbf{40.6} & 28.3 / 22.3 & \textbf{42.7} / 42.4 & \textbf{33.8} / \textbf{33.6} \\ 
  \hline
\end{tabular}
\caption{Agreement with human ratings on the WMT19 Metrics Shared Task. The metrics are Kendall Tau ($\tau$) and WMT's Direct Assessment metrics divided by 100. All the values reported for Yisi1\_SRL and ESIM fall within 0.2 percentage of the official WMT results.}
\label{table:wmt19}
\end{table*}

\section{Experiments}
\label{sec:experiments}
In this section, we report our experimental results for two tasks, translation and data-to-text. First, we benchmark \BLEURT{} against existing text generation metrics on the last 3 years of the WMT Metrics Shared Task~\cite{ma2017results}. We then evaluate its robustness to quality drifts with a series of synthetic datasets based on WMT17. We test \BLEURT{}'s ability to adapt to different tasks with the WebNLG 2017 Challenge Dataset~\cite{gardent2017webnlg}. Finally, we measure the contribution of each pre-training task with ablation experiments.

\paragraph{Our Models:} Unless specified otherwise, all \BLEURT{} models are trained in three steps: regular BERT pre-training~\cite{devlin2018bert}, pre-training on synthetic data (as explained in Section~\ref{sec:pretraining}), and fine-tuning on task-specific ratings (translation and/or data-to-text).
We experiment with two versions of \BLEURT{}, \BLEURTsys{} and \BLEURTbase, respectively based on BERT-Large (24 layers, 1024 hidden units, 16 heads) and BERT-Base (12 layers, 768 hidden units, 12 heads)~\cite{devlin2018bert}, both uncased. We use batch size 32, learning rate 1e-5, and 800,000 steps for pre-training and 40,000 steps for fine-tuning. We provide the full detail of our training setup in the Appendix.

\subsection{WMT Metrics Shared Task}
\label{sec:wmt}

\paragraph{Datasets and Metrics:} We use years 2017 to 2019 of the WMT Metrics Shared Task, to-English language pairs. For each year, we used the official WMT test set, which include several thousand pairs of sentences with human ratings from the news domain. The training sets contain 5,360, 9,492, and 147,691 records for each year. The test sets for years 2018 and 2019 are noisier, as reported by the organizers and shown by the overall lower correlations.

We evaluate the agreement between the automatic metrics and the human ratings. For each year, we report two metrics: Kendall's Tau $\tau$ (for consistency across experiments), and the official WMT metric for that year (for completeness). The official WMT metric is either Pearson's correlation or a robust variant of Kendall's Tau called \textsc{DARR}, described in the Appendix. All the numbers come from our own implementation of the benchmark.\footnote{The official scripts are public but they suffer from documentation and dependency issues, as shown by a \texttt{README} file in the 2019 edition which explicitly discourages using them.} Our results are globally consistent with the official results but we report small differences in 2018 and 2019, marked in the tables.

\paragraph{Models:} We experiment with four versions of \BLEURT{}: \BLEURTsys, \BLEURTbase, \BLEURTnopre{} and \BLEURTbasenopre. The first two models are based on BERT-large and BERT-base. In the latter two versions, we skip the pre-training phase and fine-tune directly on the WMT ratings. For each year of the WMT shared task, we use the test set from the previous years for training and validation. We describe our setup in further detail in the Appendix. We compare \BLEURT{} to participant data from the shared task and automatic metrics that we ran ourselves. In the former case, we use the the best-performing contestants for each year, that is, \texttt{chrF++}, \texttt{BEER}, \texttt{Meteor++}, \texttt{RUSE}, \texttt{Yisi1}, \texttt{ESIM} and \texttt{Yisi1-SRL}~\citep{mathur2019putting}. All the contestants use the same WMT training data, in addition to existing sentence or token embeddings. In the latter case, we use Moses \texttt{sentenceBLEU}, \texttt{BERTscore}~\cite{zhang2019bertscore}, and \texttt{MoverScore}~\citep{zhao2019moverscore}. For \texttt{BERTscore}, we use BERT-large uncased for fairness, and roBERTa (the recommended version) for completeness~\citep{liu2019roberta}. We run \texttt{MoverScore} on WMT 2017 using the scripts published by the authors.

\paragraph{Results:} Tables~\ref{table:wmt17}, \ref{table:wmt18}, \ref{table:wmt19} show the results. 
For years 2017 and 2018, a \BLEURT{}-based metric dominates the benchmark for each language pair (Tables~\ref{table:wmt17} and \ref{table:wmt18}). \BLEURTsys{} and \BLEURTbase{} are also competitive for year 2019: they yield the best results for all language pairs on Kendall's Tau, and they come first for 3 out of 7 pairs on DARR. As expected, \BLEURTsys{} dominates \BLEURTbase{} in the majority of cases. Pre-training consistently improves the results of \BLEURTsys{} and \BLEURTbase. We observe the largest effect on year 2017, where it adds up to 7.4 Kendall Tau  points for \BLEURTbase{} (\texttt{zh-en}). The effect is milder on years 2018 and 2019, up to 2.1 points (\texttt{tr-en}, 2018). We explain the difference by the fact that the training data used for 2017 is smaller than the  datasets used for the following years, so pre-training is likelier to help. In general pre-training yields higher returns for BERT-base than for BERT-large---in fact, \BLEURTbase{} with pre-training is often better than \BLEURTsys{} without.


\textbf{Takeaways:} Pre-training delivers consistent improvements, especially for BLEURT-base. \BLEURT{} yields state-of-the art performance for all years of the WMT Metrics Shared task.

\subsection{Robustness to Quality Drift}
\label{sec:drift}

\begin{figure}[!t]
\centering
\includegraphics[width=.75\columnwidth]{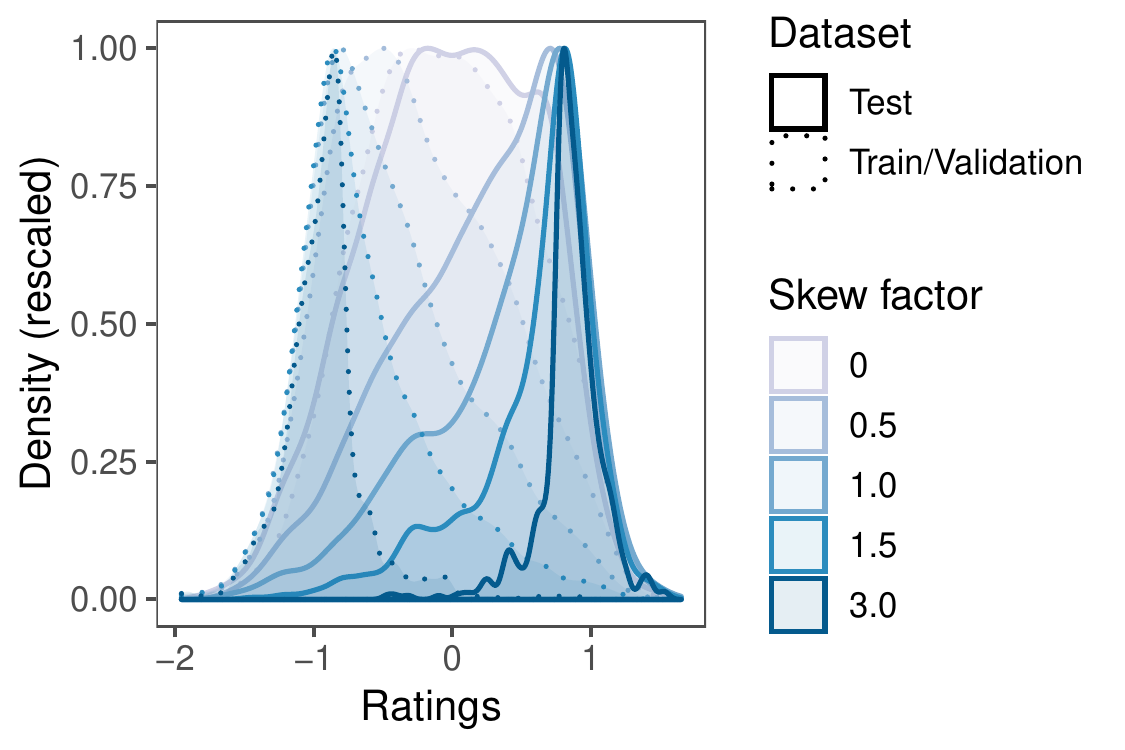}
\caption{Distribution of the human ratings in the train/validation and test datasets for different skew factors.}
\label{fig:skew}
\end{figure}

\begin{figure}[!t]
\centering
\includegraphics[width=.75\columnwidth]{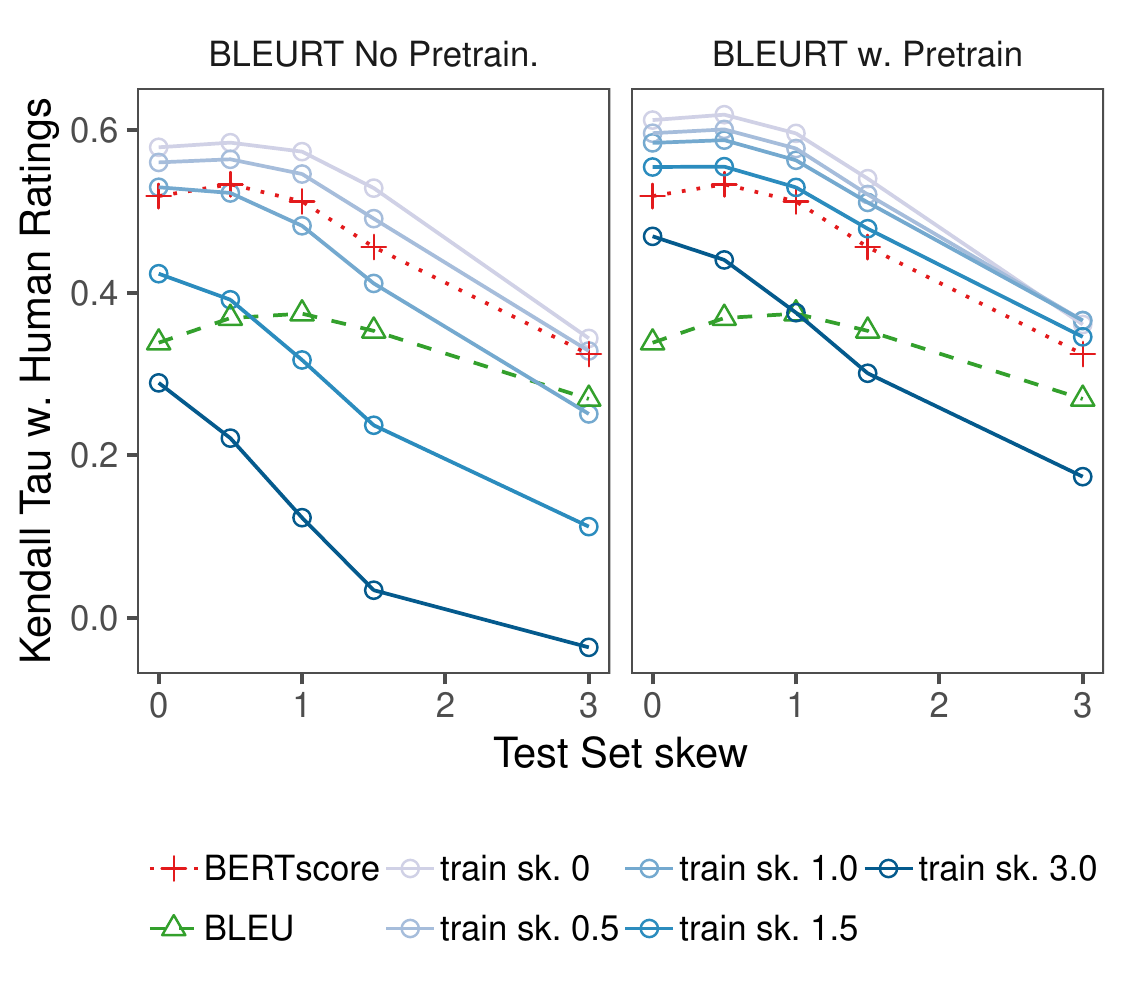}
\caption{Agreement between BLEURT and human ratings for different skew factors in train and test.}
\label{fig:extrapolation}
\end{figure}

\begin{figure*}[!t]
\centering 
\includegraphics[width=.9\textwidth]{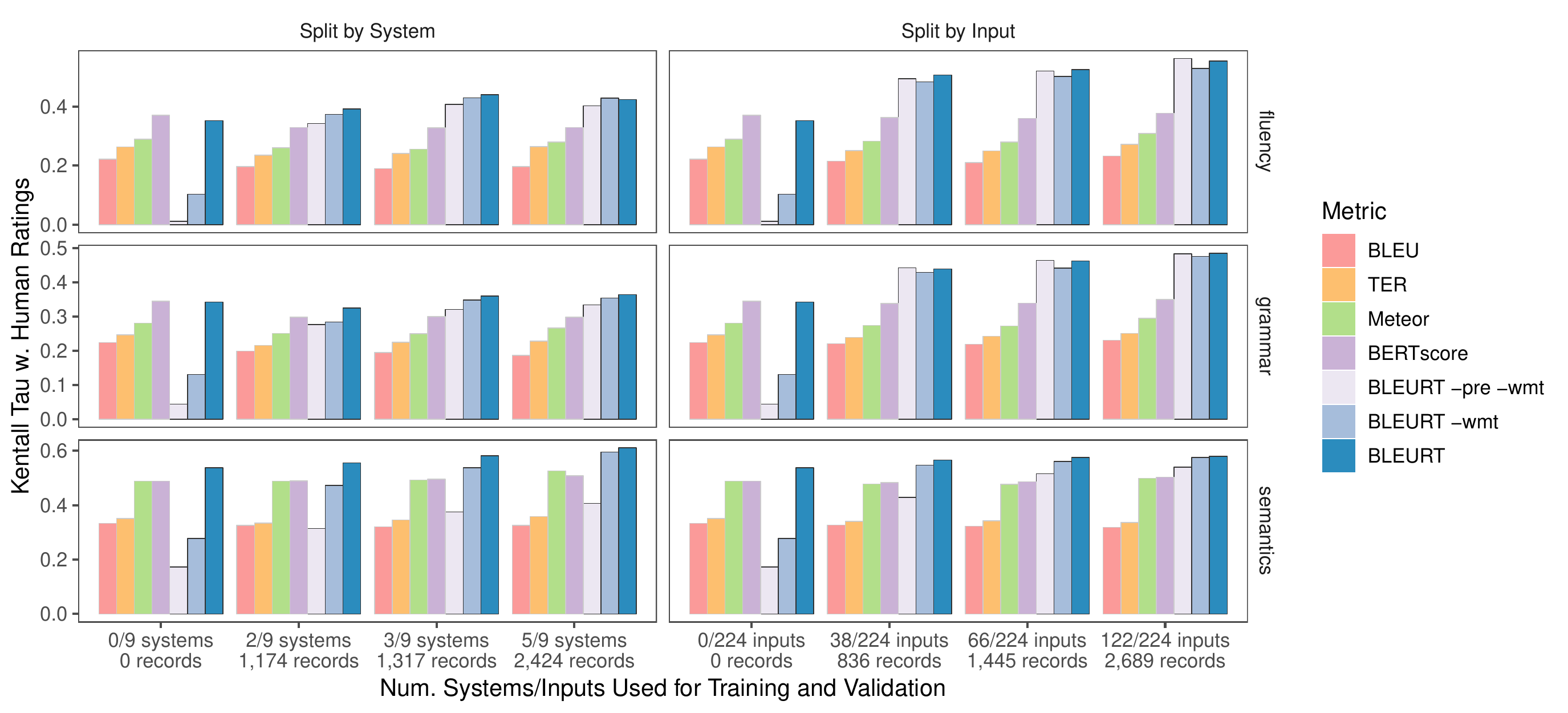}
\caption{Absolute Kendall Tau of BLEU, Meteor, and BLEURT with human judgements on the WebNLG dataset, varying the size of the data used for training and validation.}
\label{fig:webnlg}
\end{figure*}

We assess our claim that pre-training makes \BLEURT{} robust to quality drifts, by constructing a series of tasks for which it is increasingly pressured to extrapolate. All the experiments that follow are based on the WMT Metrics Shared Task 2017, because the ratings for this edition are particularly reliable.\footnote{The organizers managed to collect 15 adequacy scores for each translation, and thus the ratings are almost perfectly repeatable~\cite{ma2017results}}

\paragraph{Methodology:} We create increasingly challenging datasets by sub-sampling the records from the WMT Metrics shared task, keeping low-rated translations for training and high-rated translations for test. The key parameter is the \emph{skew factor} $\alpha$, that measures how much the training data is left-skewed and the test data is right-skewed. Figure~\ref{fig:skew} demonstrates the ratings distribution that we used in our experiments. The training data shrinks as $\alpha$ increases: in the most extreme case ($\alpha=3.0$), we use only 11.9\% of the original 5,344 training records. We give the full detail of our sampling methodology in the Appendix.

We use BLEURT with and without pre-training and we compare to Moses \texttt{sentBLEU} and \texttt{BERTscore}. We use BERT-large uncased for both \BLEURTsys{} and \texttt{BERTscore}.

\paragraph{Results:} Figure~\ref{fig:extrapolation} presents \BLEURT{}'s performance as we vary the train and test skew independently. Our first observation is that the agreements fall for all metrics as we increase the test skew. This effect was already described is the 2019 WMT Metrics report~\cite{ma2019results}. A common explanation is that the task gets more difficult as the ratings get closer---it is easier to discriminate between ``good'' and ``bad'' systems than to rank ``good'' systems. 

Training skew has a disastrous effect on \BLEURT{} without pre-training: it is below \texttt{BERTscore} for $\alpha=1.0$, and it falls under \texttt{sentBLEU} for $\alpha \geq 1.5$. Pre-trained \BLEURT{} is much more robust: the only case in which it falls under the baselines is $\alpha=3.0$, the most extreme drift, for which incorrect translations are used for train while excellent ones for test.

\paragraph{Takeaways:} Pre-training makes BLEURT significantly more robust to quality drifts.

\subsection{WebNLG Experiments}

In this section, we evaluate \BLEURT{}'s performance on three tasks from a data-to-text dataset, the WebNLG Challenge 2017~\cite{webnlg-metrics}. The aim is to assess \BLEURT{}'s capacity to adapt to new tasks with limited training data.

\paragraph{Dataset and Evaluation Tasks:} The WebNLG challenge benchmarks systems that produce natural language description of entities (e.g.,  buildings, cities, artists) from sets of 1 to 5 RDF triples. The organizers released the human assessments for 9 systems over 223 inputs, that is, 4,677 sentence pairs in total (we removed null values). Each input comes with 1 to 3 reference descriptions. The submissions are evaluated on 3 aspects: semantics, grammar, and fluency. We treat each type of rating as a separate modeling task. The data has no natural split between train and test, therefore we experiment with several schemes. We allocate 0\% to about 50\% of the data to training, and we split on both the evaluated systems or the RDF inputs in order to test different generalization regimes.


\paragraph{Systems and Baselines:} \BLEURTnoprenowmt, is a public BERT-large uncased checkpoint directly trained on the WebNLG ratings. \BLEURTnowmt was first pre-trained on synthetic data, then fine-tuned on WebNLG data. \BLEURTsys{} was trained in three steps: first on synthetic data, then on WMT data (16-18), and finally on WebNLG data. When a record comes with several references, we run BLEURT on each reference and report the highest value~\cite{zhang2019bertscore}.

We report four baselines: \texttt{BLEU}, \texttt{TER}, \texttt{Meteor}, and \texttt{BERTscore}. The first three were computed by the WebNLG competition organizers. We ran the latter one ourselves, using BERT-large uncased for a fair comparison.
 
\paragraph{Results:} Figure~\ref{fig:webnlg} presents the correlation of the metrics with human assessments as we vary the share of data allocated to training. The more pre-trained \BLEURT{} is, the quicker it adapts. The vanilla BERT approach \BLEURTnoprenowmt{} requires about a third of the WebNLG data to dominate the baselines on the majority of tasks, and it still lags behind on \emph{semantics} (split by system). In contrast, \BLEURTnowmt{} is competitive with as little as 836 records, and \BLEURT{} is comparable with \texttt{BERTscore} with zero fine-tuning.



\paragraph{Takeaways:} Thanks to pre-training, \BLEURT{} can quickly adapt to the new tasks. \BLEURT{} fine-tuned twice (first on synthetic data, then on WMT data) provides acceptable results on all tasks without training data.

\begin{figure}[t!]
\centering
\includegraphics[width=.9\columnwidth]{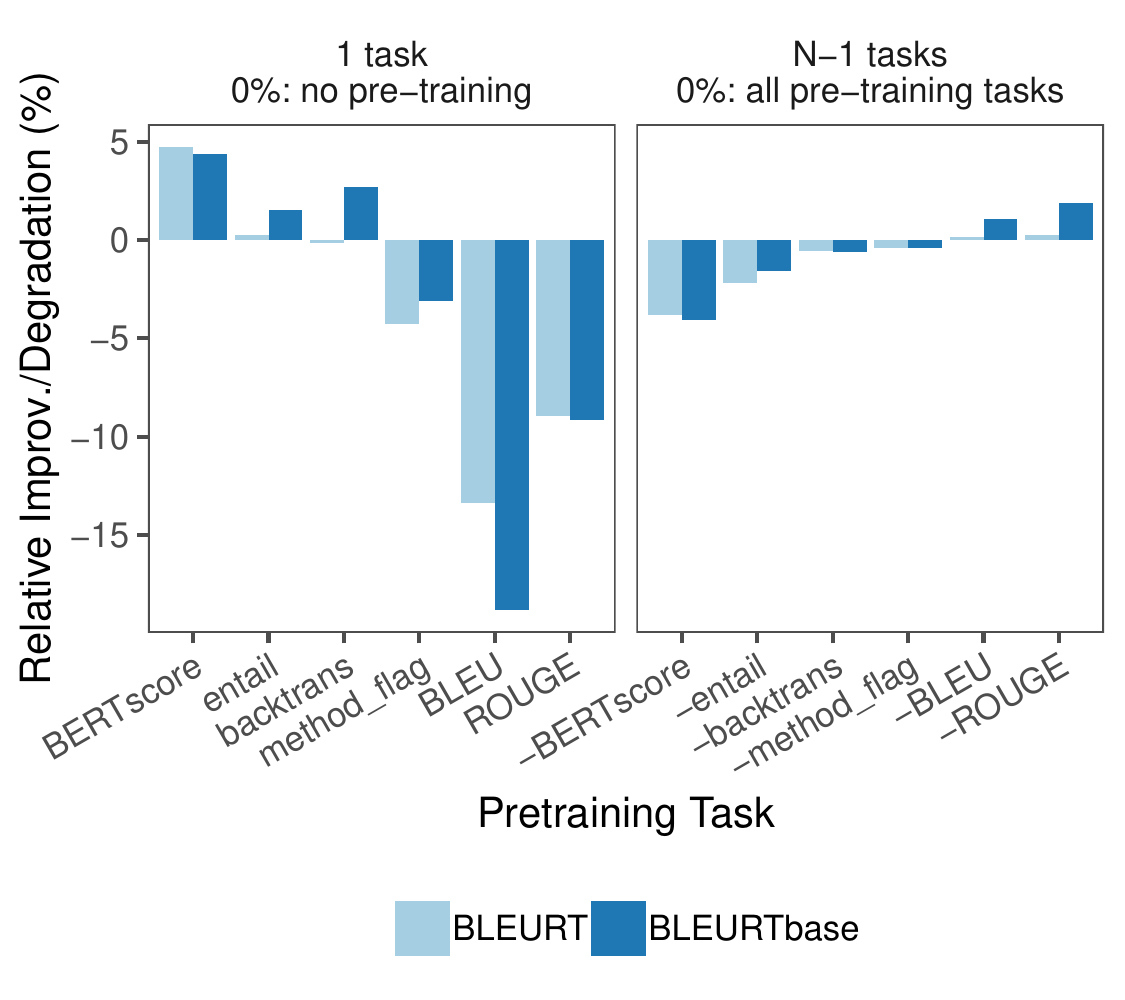}
\caption{Improvement in Kendall Tau on WMT 17 varying the pre-training tasks.}
\label{fig:pre-training-task}
\end{figure}

\subsection{Ablation Experiments}

Figure~\ref{fig:pre-training-task} presents our ablation experiments on WMT 2017, which highlight the relative importance of each pre-training task. On the left side, we compare \BLEURT{} pre-trained on a single task to \BLEURT{} without pre-training. On the right side, we compare full \BLEURT{} to \BLEURT{} pre-trained on all tasks except one. Pre-training on BERTscore, entailment, and the backtranslation scores yield improvements (symmetrically, ablating them degrades \BLEURT{}). Oppositely, BLEU and ROUGE have a negative impact. We conclude that pre-training on high quality signals helps BLEURT, but that metrics that correlate less well with human judgment may in fact harm the model.\footnote{Do those results imply that BLEU and ROUGE should be removed from future versions of \BLEURT{}? Doing so may indeed yield slight improvements on the WMT Metrics 2017 shared task. On the other hand the removal may hurt future tasks in which  BLEU or ROUGE actually correlate with human assessments. We therefore leave the question open.}

\section{Related Work}

The WMT shared metrics competition~\citep{bojar2016results,ma2018results,ma2019results} has inspired the creation of many learned metrics, some of which use regression or deep learning~\citep{stanojevic2014beer,ma2017blend,shimanaka2018ruse,chen2016enhanced, mathur2019putting}. Other metrics have been introduced, such as the recent MoverScore \cite{zhao2019moverscore} which combines contextual embeddings and Earth Mover's Distance. We provide a head-to-head comparison with the best performing of those in our experiments. Other approaches do not attempt to estimate quality directly, but use information extraction or question answering as a proxy~\citep{wiseman2017challenges,goodrich2019assessing,eyal2019question}. Those are complementary to our work.

There has been recent work that uses BERT for evaluation. BERTScore~\cite{zhang2019bertscore} proposes replacing the hard n-gram overlap of BLEU with a soft-overlap using BERT embeddings. We use it in all our experiments. Bertr~\citep{mathur2019putting} and YiSi~\cite{mathur2019putting} also make use of BERT embeddings to capture similarity. Sum-QE~\citep{xenouleassum} fine-tunes BERT for quality estimation as we describe in Section~\ref{sec:fine-tuning}. Our focus is different---we train metrics that are not only state-of-the-art in conventional \iid{} experimental setups, but also robust in the presence of scarce and out-of-distribution training data. To our knowledge no existing work has explored pre-training and extrapolation in the context of NLG.

Previous studies have used noising for referenceless evaluation~\cite{duvsek2019automatic}.
Noisy pre-training has also been proposed before for other tasks such as paraphrasing~\cite{wieting2015towards,tomar2017neural} but generally not with synthetic data. Generating synthetic data via paraphrases and perturbations has been commonly used for generating adversarial examples~\cite{jia2017adversarial,iyyer2018adversarial,BelinkovB18,ribeiro2018semantically}, an orthogonal line of research.

\section{Conclusion}

We presented \BLEURT{}, a reference-based text generation metric for English. Because the metric is trained end-to-end, \BLEURT{} can model human assessment with superior accuracy. Furthermore, pre-training makes the metrics robust particularly robust to both domain and quality drifts. Future research directions include multilingual NLG evaluation, and hybrid methods involving both humans and classifiers. 

\section*{Acknowledgments}
Thanks to Eunsol Choi, Nicholas FitzGerald, Jacob Devlin, and to the members of the Google AI Language team for the proof-reading, feedback, and suggestions. We also thank Madhavan Kidambi and Ming-Wei Chang, who implemented blank-filling with BERT.


\bibliography{main}

\begin{thebibliography}{49}
\expandafter\ifx\csname natexlab\endcsname\relax\def\natexlab#1{#1}\fi

\bibitem[{Bahdanau et~al.(2015)Bahdanau, Cho, and Bengio}]{bahdanau2014neural}
Dzmitry Bahdanau, Kyunghyun Cho, and Yoshua Bengio. 2015.
\newblock Neural machine translation by jointly learning to align and
  translate.
\newblock In \emph{Proceedings of ICLR}.

\bibitem[{Bannard and Callison-Burch(2005)}]{bannard2005paraphrasing}
Colin Bannard and Chris Callison-Burch. 2005.
\newblock Paraphrasing with bilingual parallel corpora.
\newblock In \emph{Proceedings of ACL}.

\bibitem[{Belinkov and Bisk(2018)}]{BelinkovB18}
Yonatan Belinkov and Yonatan Bisk. 2018.
\newblock Synthetic and natural noise both break neural machine translation.
\newblock In \emph{Proceedings of ICLR}.

\bibitem[{Bojar et~al.(2017)Bojar, Graham, and Kamran}]{ma2017results}
Ond{\v{r}}ej Bojar, Yvette Graham, and Amir Kamran. 2017.
\newblock Results of the wmt17 metrics shared task.
\newblock In \emph{Proceedings of WMT}.

\bibitem[{Bojar et~al.(2016)Bojar, Graham, Kamran, and
  Stanojevi{\'c}}]{bojar2016results}
Ond{\v{r}}ej Bojar, Yvette Graham, Amir Kamran, and Milo{\v{s}} Stanojevi{\'c}.
  2016.
\newblock Results of the wmt16 metrics shared task.
\newblock In \emph{Proceedings of WMT}.

\bibitem[{Bowman et~al.(2015)Bowman, Angeli, Potts, and
  Manning}]{bowman2015large}
Samuel~R Bowman, Gabor Angeli, Christopher Potts, and Christopher~D Manning.
  2015.
\newblock A large annotated corpus for learning natural language inference.
\newblock \emph{Proceedings of EMNLP}.

\bibitem[{Chaganty et~al.(2018)Chaganty, Mussman, and
  Liang}]{chaganty2018price}
Arun~Tejasvi Chaganty, Stephen Mussman, and Percy Liang. 2018.
\newblock The price of debiasing automatic metrics in natural language
  evaluation.
\newblock \emph{Proceedings of ACL}.

\bibitem[{Chen et~al.(2017)Chen, Zhu, Ling, Wei, Jiang, and
  Inkpen}]{chen2016enhanced}
Qian Chen, Xiaodan Zhu, Zhenhua Ling, Si~Wei, Hui Jiang, and Diana Inkpen.
  2017.
\newblock Enhanced lstm for natural language inference.
\newblock \emph{Proceedings of ACL}.

\bibitem[{Chopra et~al.(2016)Chopra, Auli, and Rush}]{chopra2016abstractive}
Sumit Chopra, Michael Auli, and Alexander~M Rush. 2016.
\newblock Abstractive sentence summarization with attentive recurrent neural
  networks.
\newblock In \emph{Proceedings of NAACL HLT}.

\bibitem[{Devlin et~al.(2019)Devlin, Chang, Lee, and
  Toutanova}]{devlin2018bert}
Jacob Devlin, Ming-Wei Chang, Kenton Lee, and Kristina Toutanova. 2019.
\newblock Bert: Pre-training of deep bidirectional transformers for language
  understanding.
\newblock In \emph{Proceedings of NAACL HLT}.

\bibitem[{Du{\v{s}}ek et~al.(2019)Du{\v{s}}ek, Sevegnani, Konstas, and
  Rieser}]{duvsek2019automatic}
Ond{\v{r}}ej Du{\v{s}}ek, Karin Sevegnani, Ioannis Konstas, and Verena Rieser.
  2019.
\newblock Automatic quality estimation for natural language generation: Ranting
  (jointly rating and ranking).
\newblock \emph{Proceedings of INLG}.

\bibitem[{Eyal et~al.(2019)Eyal, Baumel, and Elhadad}]{eyal2019question}
Matan Eyal, Tal Baumel, and Michael Elhadad. 2019.
\newblock Question answering as an automatic evaluation metric for news article
  summarization.
\newblock In \emph{Proceedings of NAACL HLT}.

\bibitem[{Fang et~al.(2015)Fang, Gupta, Iandola, Srivastava, Deng, Doll{\'a}r,
  Gao, He, Mitchell, Platt et~al.}]{fang2015captions}
Hao Fang, Saurabh Gupta, Forrest Iandola, Rupesh~K Srivastava, Li~Deng, Piotr
  Doll{\'a}r, Jianfeng Gao, Xiaodong He, Margaret Mitchell, John~C Platt,
  et~al. 2015.
\newblock From captions to visual concepts and back.
\newblock In \emph{Proceedings of CVPR}.

\bibitem[{Ganitkevitch et~al.(2013)Ganitkevitch, Van~Durme, and
  Callison-Burch}]{ganitkevitch2013ppdb}
Juri Ganitkevitch, Benjamin Van~Durme, and Chris Callison-Burch. 2013.
\newblock Ppdb: The paraphrase database.
\newblock In \emph{Proceedings NAACL HLT}.

\bibitem[{Gardent et~al.(2017)Gardent, Shimorina, Narayan, and
  Perez-Beltrachini}]{gardent2017webnlg}
Claire Gardent, Anastasia Shimorina, Shashi Narayan, and Laura
  Perez-Beltrachini. 2017.
\newblock The webnlg challenge: Generating text from rdf data.
\newblock In \emph{Proceedings of INLG}.

\bibitem[{Goodrich et~al.(2019)Goodrich, Saleh, Liu, and
  Rao}]{goodrich2019assessing}
Ben Goodrich, Mohammad~Ahmad Saleh, Peter Liu, and Vinay Rao. 2019.
\newblock Assessing the factual accuracy of text generation.
\newblock In \emph{Proceedings of ACM SIGKDD}.

\bibitem[{Iyyer et~al.(2018)Iyyer, Wieting, Gimpel, and
  Zettlemoyer}]{iyyer2018adversarial}
Mohit Iyyer, John Wieting, Kevin Gimpel, and Luke Zettlemoyer. 2018.
\newblock Adversarial example generation with syntactically controlled
  paraphrase networks.
\newblock \emph{Proceedings of NAACL HLT}.

\bibitem[{Jia and Liang(2017)}]{jia2017adversarial}
Robin Jia and Percy Liang. 2017.
\newblock Adversarial examples for evaluating reading comprehension systems.
\newblock \emph{Proceedings of EMNLP}.

\bibitem[{Koehn(2009)}]{koehn2009statistical}
Philipp Koehn. 2009.
\newblock \emph{Statistical machine translation}.
\newblock Cambridge University Press.

\bibitem[{Kukich(1983)}]{kukich1983design}
Karen Kukich. 1983.
\newblock Design of a knowledge-based report generator.
\newblock In \emph{Proceedings of ACL}.

\bibitem[{Lin(2004)}]{lin2004rouge}
Chin-Yew Lin. 2004.
\newblock Rouge: A package for automatic evaluation of summaries.
\newblock In \emph{Workshop on Text Summarization Branches Out}.

\bibitem[{Liu et~al.(2016)Liu, Lowe, Serban, Noseworthy, Charlin, and
  Pineau}]{liu2016not}
Chia-Wei Liu, Ryan Lowe, Iulian~V Serban, Michael Noseworthy, Laurent Charlin,
  and Joelle Pineau. 2016.
\newblock How not to evaluate your dialogue system: An empirical study of
  unsupervised evaluation metrics for dialogue response generation.
\newblock \emph{Proceedings of EMNLP}.

\bibitem[{Liu et~al.(2019)Liu, Ott, Goyal, Du, Joshi, Chen, Levy, Lewis,
  Zettlemoyer, and Stoyanov}]{liu2019roberta}
Yinhan Liu, Myle Ott, Naman Goyal, Jingfei Du, Mandar Joshi, Danqi Chen, Omer
  Levy, Mike Lewis, Luke Zettlemoyer, and Veselin Stoyanov. 2019.
\newblock Roberta: A robustly optimized bert pretraining approach.
\newblock \emph{arXiv:1907.11692}.

\bibitem[{Ma et~al.(2018)Ma, Bojar, and Graham}]{ma2018results}
Qingsong Ma, Ond{\v{r}}ej Bojar, and Yvette Graham. 2018.
\newblock Results of the wmt18 metrics shared task: Both characters and
  embeddings achieve good performance.
\newblock In \emph{Proceedings of WMT}.

\bibitem[{Ma et~al.(2017)Ma, Graham, Wang, and Liu}]{ma2017blend}
Qingsong Ma, Yvette Graham, Shugen Wang, and Qun Liu. 2017.
\newblock Blend: a novel combined mt metric based on direct
  assessment--casict-dcu submission to wmt17 metrics task.
\newblock In \emph{Proceedings of WMT}.

\bibitem[{Ma et~al.(2019)Ma, Wei, Bojar, and Graham}]{ma2019results}
Qingsong Ma, Johnny Wei, Ond{\v{r}}ej Bojar, and Yvette Graham. 2019.
\newblock Results of the wmt19 metrics shared task: Segment-level and strong mt
  systems pose big challenges.
\newblock In \emph{Proceedings of WMT}.

\bibitem[{Mani(1999)}]{mani1999advances}
Inderjeet Mani. 1999.
\newblock \emph{Advances in automatic text summarization}.
\newblock MIT press.

\bibitem[{Mathur et~al.(2019)Mathur, Baldwin, and Cohn}]{mathur2019putting}
Nitika Mathur, Timothy Baldwin, and Trevor Cohn. 2019.
\newblock Putting evaluation in context: Contextual embeddings improve machine
  translation evaluation.
\newblock In \emph{Proceedings of ACL}.

\bibitem[{McKeown(1992)}]{mckeown1992text}
Kathleen McKeown. 1992.
\newblock \emph{Text generation}.
\newblock Cambridge University Press.

\bibitem[{Novikova et~al.(2017)Novikova, Du{\v{s}}ek, Curry, and
  Rieser}]{novikova2017we}
Jekaterina Novikova, Ond{\v{r}}ej Du{\v{s}}ek, Amanda~Cercas Curry, and Verena
  Rieser. 2017.
\newblock Why we need new evaluation metrics for nlg.
\newblock \emph{Proceedings of EMNLP}.

\bibitem[{Papineni et~al.(2002)Papineni, Roukos, Ward, and
  Zhu}]{papineni2002bleu}
Kishore Papineni, Salim Roukos, Todd Ward, and Wei-Jing Zhu. 2002.
\newblock Bleu: a method for automatic evaluation of machine translation.
\newblock In \emph{Proceedings of ACL}.

\bibitem[{Ribeiro et~al.(2018)Ribeiro, Singh, and
  Guestrin}]{ribeiro2018semantically}
Marco~Tulio Ribeiro, Sameer Singh, and Carlos Guestrin. 2018.
\newblock Semantically equivalent adversarial rules for debugging nlp models.
\newblock In \emph{Proceedings of ACL}.

\bibitem[{Sennrich et~al.(2016)Sennrich, Haddow, and
  Birch}]{sennrich2015improving}
Rico Sennrich, Barry Haddow, and Alexandra Birch. 2016.
\newblock Improving neural machine translation models with monolingual data.
\newblock \emph{Proceedings of ACL}.

\bibitem[{Shimanaka et~al.(2018)Shimanaka, Kajiwara, and
  Komachi}]{shimanaka2018ruse}
Hiroki Shimanaka, Tomoyuki Kajiwara, and Mamoru Komachi. 2018.
\newblock Ruse: Regressor using sentence embeddings for automatic machine
  translation evaluation.
\newblock In \emph{Proceedings of WMT}.

\bibitem[{Shimorina et~al.(2019)Shimorina, Gardent, Narayan, and
  Perez-Beltrachini}]{webnlg-metrics}
Anastasia Shimorina, Claire Gardent, Shashi Narayan, and Laura
  Perez-Beltrachini. 2019.
\newblock Webnlg challenge: Human evaluation results.
\newblock Technical report.

\bibitem[{Smith and Hipp(1994)}]{smith1994spoken}
Ronnie~W Smith and D~Richard Hipp. 1994.
\newblock \emph{Spoken natural language dialog systems: A practical approach}.
\newblock Oxford University Press.

\bibitem[{Stanojevic and Sima’an(2014)}]{stanojevic2014beer}
Milos Stanojevic and Khalil Sima’an. 2014.
\newblock Beer: Better evaluation as ranking.
\newblock In \emph{Proceedings of WMT}.

\bibitem[{Sutskever et~al.(2014)Sutskever, Vinyals, and
  Le}]{sutskever2014sequence}
Ilya Sutskever, Oriol Vinyals, and Quoc~V Le. 2014.
\newblock Sequence to sequence learning with neural networks.
\newblock In \emph{Proceedings of NIPS}.

\bibitem[{Tian et~al.(2019)Tian, Narayan, Sellam, and
  Parikh}]{tian2019sticking}
Ran Tian, Shashi Narayan, Thibault Sellam, and Ankur~P Parikh. 2019.
\newblock Sticking to the facts: Confident decoding for faithful data-to-text
  generation.
\newblock \emph{arXiv:1910.08684}.

\bibitem[{Tomar et~al.(2017)Tomar, Duque, T{\"a}ckstr{\"o}m, Uszkoreit, and
  Das}]{tomar2017neural}
Gaurav~Singh Tomar, Thyago Duque, Oscar T{\"a}ckstr{\"o}m, Jakob Uszkoreit, and
  Dipanjan Das. 2017.
\newblock Neural paraphrase identification of questions with noisy pretraining.
\newblock \emph{Proceedings of the First Workshop on Subword and Character
  Level Models in NLP}.

\bibitem[{Vaswani et~al.(2017)Vaswani, Shazeer, Parmar, Uszkoreit, Jones,
  Gomez, Kaiser, and Polosukhin}]{vaswani2017attention}
Ashish Vaswani, Noam Shazeer, Niki Parmar, Jakob Uszkoreit, Llion Jones,
  Aidan~N Gomez, {\L}ukasz Kaiser, and Illia Polosukhin. 2017.
\newblock Attention is all you need.
\newblock In \emph{Proceedings of NIPS}.

\bibitem[{Vinyals and Le(2015)}]{vinyals2015neural}
Oriol Vinyals and Quoc Le. 2015.
\newblock A neural conversational model.
\newblock \emph{Proceedings of ICML}.

\bibitem[{Wang et~al.(2019)Wang, Singh, Michael, Hill, Levy, and
  Bowman}]{wang2018glue}
Alex Wang, Amanpreet Singh, Julian Michael, Felix Hill, Omer Levy, and Samuel~R
  Bowman. 2019.
\newblock Glue: A multi-task benchmark and analysis platform for natural
  language understanding.
\newblock \emph{Proceedings of ICLR}.

\bibitem[{Wieting et~al.(2016)Wieting, Bansal, Gimpel, and
  Livescu}]{wieting2015towards}
John Wieting, Mohit Bansal, Kevin Gimpel, and Karen Livescu. 2016.
\newblock Towards universal paraphrastic sentence embeddings.
\newblock \emph{Proceedings of ICLR}.

\bibitem[{Williams et~al.(2018)Williams, Nangia, and
  Bowman}]{williams2017broad}
Adina Williams, Nikita Nangia, and Samuel~R Bowman. 2018.
\newblock A broad-coverage challenge corpus for sentence understanding through
  inference.
\newblock \emph{Proceedings of NAACL HLT}.

\bibitem[{Wiseman et~al.(2017)Wiseman, Shieber, and
  Rush}]{wiseman2017challenges}
Sam Wiseman, Stuart~M Shieber, and Alexander~M Rush. 2017.
\newblock Challenges in data-to-document generation.
\newblock \emph{Proceedings of EMNLP}.

\bibitem[{Xenouleas et~al.(2019)Xenouleas, Malakasiotis, Apidianaki, and
  Androutsopoulos}]{xenouleassum}
Stratos Xenouleas, Prodromos Malakasiotis, Marianna Apidianaki, and Ion
  Androutsopoulos. 2019.
\newblock Sum-qe: a bert-based summary quality estimation model supplementary
  material.
\newblock In \emph{Proceedings of EMNLP}.

\bibitem[{Zhang et~al.(2020)Zhang, Kishore, Wu, Weinberger, and
  Artzi}]{zhang2019bertscore}
Tianyi Zhang, Varsha Kishore, Felix Wu, Kilian~Q Weinberger, and Yoav Artzi.
  2020.
\newblock Bertscore: Evaluating text generation with bert.
\newblock \emph{Proceedings of ICLR}.

\bibitem[{Zhao et~al.(2019)Zhao, Peyrard, Liu, Gao, Meyer, and
  Eger}]{zhao2019moverscore}
Wei Zhao, Maxime Peyrard, Fei Liu, Yang Gao, Christian~M Meyer, and Steffen
  Eger. 2019.
\newblock Moverscore: Text generation evaluating with contextualized embeddings
  and earth mover distance.
\newblock \emph{Proceedings of EMNLP}.

\end{thebibliography}
\bibliographystyle{acl_natbib}

\appendix

\section{Implementation Details of the Pre-Training Phase}

This section provides implementation details for some of the pre-training techniques described in the main paper.

\subsection{Data Generation}

\paragraph{Random Masking:} We use two masking strategies. The first strategy samples random words in the sentence and it replaces them with masks (one for each token). Thus, the masks are scattered across the sentence. The second strategy creates contiguous sequences: it samples a start position $s$, a length $l$ (uniformly distributed), and it masks all the tokens spanned by words between positions $s$ and $s+l$. In both cases, we use up to 15 masks per sentence. Instead of running the language model once and picking the most likely token at each position, we use beam search (the beam size 8 by default). This enforces consistency and avoids repeated sequences, e.g., ``\texttt{,,,}''.

\paragraph{Backtranslation:} Consider English and French. Given a forward translation model $P_{\en \rightarrow \fr}(z_{\fr} | z_{\en})$ and backward translation model $P_{\fr \rightarrow \en}(z_{\en} | z_{\fr})$, we generate $\tilde{\vz}$ as follows:
\begin{align*}
\tilde{\vz} = \argmax_{z_{\en}} \left (P_{\fr \rightarrow \en}(z_{\en} | z_{\fr}^\ast) \right )
\end{align*}
where $z_{\fr}^\ast = \argmax_{z_{\fr}} \left ( P_{\fr \rightarrow \en}(z_{\fr} | z ) \right )$.
For the translations, we use a Transformer model~\citep{vaswani2017attention}, trained on English-German with the \texttt{tensor2tensor} framework.\footnote{\url{https://github.com/tensorflow/tensor2tensor}}

\paragraph{Word dropping:} Given a synthetic example $(\vz, \tilde{\vz})$ we generate a pair $(\vz, \tilde{\vz}')$, by randomly dropping words from $\tilde{\vz}$. We draw the number of words to drop uniformly, up to the length of the sentence. We apply this transformation on about 30\% of the data generated with the previous method.

\subsection{Pre-Training Tasks}

We now provide additional details on the signals we used for pre-training.

\paragraph{Automatic Metrics:} As shown in the table, we use three types of signals: BLEU, ROUGE, and BERTscore. For BLEU, we used the original Moses \textsc{sentenceBLEU}\footnote{\url{https://github.com/moses-smt/mosesdecoder/blob/master/mert/sentence-bleu.cpp}} implementation, using the Moses tokenizer and the default parameters. For ROUGE, we used the \texttt{seq2seq} implementation of \texttt{ROUGE-N}.\footnote{\url{https://github.com/google/seq2seq/blob/master/seq2seq/metrics/rouge.py}} We used a custom implementation of \textsc{BERTscore}, based on BERT-large uncased. ROUGE and BERTscore return three scores: precision, recall, and F-score. We use all three quantities.

\paragraph{Backtranslation Likelihood:}
We compute all the losses using custom Transformer model~\citep{vaswani2017attention}, trained on two language pairs (English-French and English-German) with the \texttt{tensor2tensor} framework.

\paragraph{Normalization:} All the regression labels are normalized before training. 

\subsection{Modeling}

\paragraph{Setting the weights of the pre-training tasks:} 
We set the weights $\gamma_k$ with grid search, optimizing \BLEURT{}'s performance on WMT 17's validation set. To reduce the size of the grid, we make groups of pre-training tasks that share the same weights: $(\bm{\tau}_{\text{BLEU}}, \bm{\tau}_{\text{ROUGE}}, \bm{\tau}_{\text{BERTscore}})$, $(\bm{\tau}_{\text{en-fr}, z \mid \tilde{z}}, \bm{\tau}_{\text{en-fr}, \tilde{z} \mid z}, \bm{\tau}_{\text{en-de}, z \mid \tilde{z}}, \bm{\tau}_{\text{en-de}, \tilde{z} \mid z})$, and $(\bm{\tau}_{\text{entail}}, \bm{\tau}_{\text{backtran\_flag}})$.

\section{Experiments--Supplementary Material}

\subsection{Training Setup for All Experiments} 
 We user BERT's public checkpoints\footnote{\url{https://github.com/google-research/bert}} with Adam (the default optimizer), learning rate 1e-5, and batch size 32. Unless specified otherwise, we use 800,00 training steps for pre-training and 40,000 steps for fine-tuning. We run training and evaluation in parallel: we run the evaluation every 1,500 steps and store the checkpoint that performs best on a held-out validation set (more details on the data splits and our choice of metrics in the following sections). We use Google Cloud TPUs v2 for learning, and Nvidia Tesla V100 accelerators for evaluation and test. Our code uses Tensorflow 1.15 and Python 2.7.

\subsection{WMT Metric Shared Task}

\paragraph{Metrics.} The metrics used to compare the evaluation systems vary across the years. The organizers use Pearson's correlation on standardized human judgments across all segments in 2017, and a custom variant of Kendall's Tau named ``DARR'' on raw human judgments in 2018 and 2019. The latter metrics operates as follows. The organizers gather all the translations for the same reference segment, they enumerate all the possible pairs $(\text{translation}_1, \text{translation}_2)$, and they discard all the pairs which have a ``similar'' score (less than 25 points away on a 100 points scale). For each remaining pair, they then determine which translation is the best according both human judgment and the candidate metric. Let $|\text{Concordant}|$ be the number of pairs on which the NLG metrics agree and $|\text{Discordant}|$ be those on which they disagree, then the score is computed as follows:
$$
\frac{|\text{Concordant}| - |\text{Discordant}|}{|\text{Concordant}| + |\text{Discordant}|}
$$
The idea behind the 25 points filter is to make the evaluation more robust, since the judgments collected for WMT 2018 and 2019 are noisy. Kendall's Tau is identical, but it does not use the filter.

\paragraph{Training setup.} To separate training and validation data, we set aside a fixed ratio of records in such a way that there is no ``leak'' between the datasets (i.e., train and validation records that share the same source). We use 10\% of the data for validation for years 2017 and 2018, and 5\% for year 2019. We report results for the models that yield the highest Kendall Tau across all records on validation data.  The weights associated to each pretraining task (see our Modeling section) are set with grid search, using the train/validation setup of WMT 2017.

\paragraph{Baselines.} we use three metrics: the Moses implementation of \texttt{sentenceBLEU},\footnote{\url{https://github.com/moses-smt/mosesdecoder/blob/master/mert/sentence-bleu.cpp}} \texttt{BERTscore},\footnote{\url{https://github.com/Tiiiger/bert_score}} and \texttt{MoverScore},\footnote{\url{https://github.com/AIPHES/emnlp19-moverscore}} which are all available online.  We run the Moses tokenizer on the reference and candidate segments before computing \texttt{sentenceBLEU}.

\begin{figure}[t!]
\centering
\includegraphics[width=.9\columnwidth]{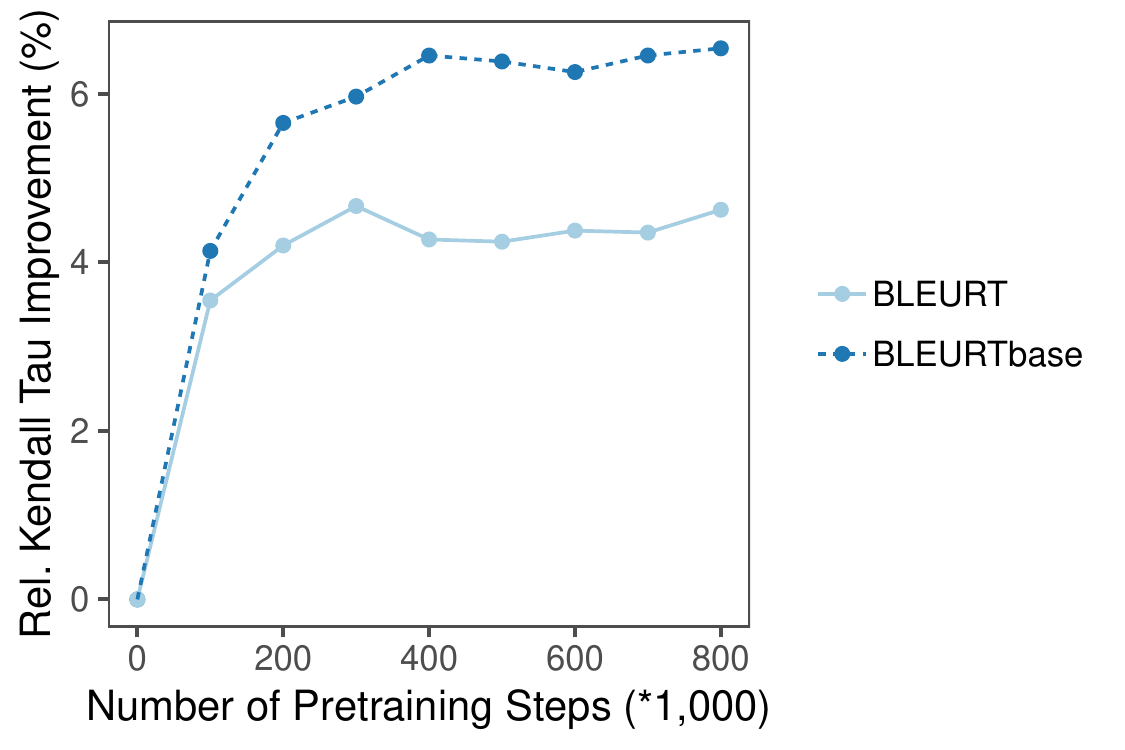}
\caption{Improvement in Kendall Tau accuracy on all language pairs of the WMT Metrics Shared Task 2017, varying the number of pre-training steps. 0 steps corresponds to 0.555 Kendall Tau for BLEURTbase and 0.580 for BLEURT.}
\label{fig:pre-training-duration}
\end{figure}

\subsection{Robustness to Quality Drift}

\paragraph{Data Re-sampling Methodology:} We sample the training and test separately, as follows. We split the data in 10 bins of equal size. We then sample each record in the dataset with probabilities $\frac{1}{B^\alpha}$  and $\frac{1}{(11-B)^\alpha}$ for train and test respectively, where $B$ is the bin index of the record between 1 and 10, and $\alpha$ is a predefined \emph{skew factor}. The skew factor $\alpha$ controls the drift: a value of 0 has no effect (the ratings are centered around 0), and value of 3.0 yields extreme differences. Note that the sizes of the datasets decrease as $\alpha$ increases: we use 50.7\%, 30.3\%, 20.4\%, and 11.9\% of the original 5,344 training records for $\alpha=0.5$, $1.0$, $1.5$, and $3.0$ respectively.

\subsection{Ablation Experiment--How Much Pre-Training Time is Necessary?}
To understand the relationship between pre-training time and downstream accuracy, we pre-train several versions of \texttt{BLEURT} and we fine-tune them on WMT17 data, varying the number of pre-training steps. Figure~\ref{fig:pre-training-duration} presents the results. Most gains are obtained during the first 400,000 steps, that is, after about 2 epochs over our synthetic dataset.



\end{document}